# Improved Repeatability Measures for Evaluating the Performance of Feature Detectors

S. Ehsan, N. Kanwal, A. F. Clark and K.D. McDonald-Maier

The most frequently employed measure for performance characterization of local feature detectors is repeatability, but it has been observed that this does not necessarily mirror actual performance. In this letter, improved repeatability formulations are presented which correlate much better with the true performance of feature detectors. Comparative results for several state-of-the-art feature detectors are presented using these measures; it is found that Hessian-based detectors are generally superior at identifying features when images are subject to various geometric and photometric transformations.

*Introduction*: The extraction of image features that are reasonably independent of scale, orientation and photometrical changes has long been an aim of the vision research community. The last decade has seen the development of a number of such operators, the best-known of which is SIFT [1-2] that incorporates both a detector and a descriptor. The evaluation of the performances of these detectors under various geometric and photometric transformations has become important, in order to identify their strengths and shortcomings for a range of vision applications [3]. Several approaches have been used for evaluating the performances of interest point detectors, including ground-truth verification, localization accuracy and theoretical analysis; however, the most widely employed measure is repeatability rate [1]. Proposed in [3], repeatability rate is defined as the ratio of the number of points repeated in the overlapping region of two images to the total number of detected points. An interest point is considered 'repeated' if its 2-D projection in the other image using planar homography lies within a neighborhood of size ε of an interest point detected in the other image.

Since these feature detectors identify interest points at different scales, measuring the 2-D distance between interest points detected at different scales, to decide whether they are repeatable or not, may lead to inaccurate results. A more refined definition of repeatability is presented in [4], which also considers the overlap of scale-dependent regions centered in the interest points. In this letter, we highlight the limitations of this definition of repeatability and propose alternatives that provide results which indicate the effect of various transformations reliably and are more consistent with the actual performance of detectors. Here, actual performance means the true matches which are obtained using ground-truth homography after descriptor based matching of detected points. Finally, a comparative analysis of interest point detectors using the various formulations is presented.

*Limitations of Repeatability*: Repeatability is conventionally defined as in [4]:

$$Repeatability = \frac{\text{Total number of repeated points}}{\min(\text{points detected in image 1}, \text{points detected in image 2})} \quad (1)$$

Despite being popular, it has been remarked that "repeatability does not guarantee high performance" [1]. Some limitations of repeatability as defined in [4] are:

I. The repeatability rate partially reflects the effect of various geometric and photometric transformations as it considers the *minimum* number of interest points detected in either of the two images.

II. It is not always possible to predict the effect of a specific transformation on the number of corresponding points from the value of repeatability.

III. The reference image is not fixed when evaluating the performance of a detector for a specific dataset.

IV. Repeatability does not always reflect the effect of transformation on the number of true matched points, *i.e.* the true performance.

To overcome the above-mentioned shortcomings, we present two alternative definitions of repeatability that are more consistent with the actual performance of feature detectors. Both use the same scale-dependent regions as [4]. The first of these is appropriate for applications that involve image sequences, while the second is more suited to applications involving pairs of images (*e.g.,* computational stereo).

*Criterion 1*: Unlike the definition in [4], the sequence of images is not ignored when determining the effect of various photometric and geometric transformations; the first image in the sequence is considered as the 'reference' in all cases. We also take into account only those interest points that lie in the common part of the two images and define an interest point as 'repeatable' if ε < 1.5 pixels and the overlap error between scale-dependent regions centered in the two interest points, defined as:

$$Overlap\ error = 1 - \frac{\mu_a \cap (A^T \mu_b A)}{(\mu_a \cup A^T \mu_b A)} \quad (2)$$

is less than 40%, as in [4], where $\mu_a$ and $\mu_b$ are the regions defined by $x^T \mu x = 1$ and A is the homography between the two images. The numerator of the fractional part in (2) represents the intersection whereas the denominator represents the union of these regions. However, as opposed to [4], which uses the minimum of the number of interest points detected in the two images, we define the repeatability rate as:

$$Criterion\ 1 = \frac{N_{rep}}{N_{ref}} \quad (3)$$

where $N_{rep}$ is the total number of repeated points and $N_{ref}$ is the total number of interest points in the common part of reference image.

*Criterion 2:* This criterion follows the same framework as described above but employs a symmetric approach for the computation of repeatability rate:

$$Criterion\ 2\ = \frac{2*N_{rep}}{N_{ref}+ N_{test}} \quad (4)$$

Where $N_{rep}$ is the number of repeated interest points, $N_{ref}$ and $N_{test}$ are the number of interest points detected in the common part of scene in reference and test image respectively.

*Results:* Repeatability values were computed for the widely-used Oxford datasets [5] using the original definition of repeatability and the two criteria defined above. Results were obtained for six state-of-the-art feature detectors, namely SIFT, SURF, Harris-Laplace, Hessian-Laplace, Harris-Affine and Hessian-Affine, using their original implementations with default parameters [2, 4, 6-7]. For all detectors, the number of true matches was also calculated for every image pair using ground-truth homography after descriptor based matching of detected points. As an example, the repeatability values and the numbers of true matches obtained for the Bark dataset (which shows variation in camera zoom and rotation) [5] with the Hessian-Laplace detector are shown in Fig. 1. (Note that repeatability values should be read from the left ordinate axis and the number of true matches from the right ordinate axis). It is evident that, in contrast to the original definition of repeatability [4], both proposed repeatability formulations vary in close agreement with the trend of number of true matches *(i.e.,* the actual performance).

To measure how well the three repeatability curves agree with the number of true matches, Pearson's correlation coefficient, r, is used. Correlation coefficient values with corresponding p-values for the SURF detector (Fast-Hessian) are given as an example in Table 1; note that a p-value gives the probability that the corresponding correlation value is *incorrect*. These results demonstrate the high reliability of repeatability values obtained using the proposed definitions. For all combinations of the six state-of-the-art feature detectors and eight datasets [5], the mean value of the correlation coefficient is: 0.844 with standard deviation 0.287 for the original criterion, 0.970 ± 0.046 for criterion 1, and 0.968 ± 0.033 for criterion 2.

Finally, a comparative analysis of six state-of-the-art detectors was carried out using the repeatability criteria defined above. Fig. 2 shows the results for the Bikes dataset (showing variation in the amount of blurring) [5] using criterion 2. A summary of the results for all Oxford datasets [5] is given in Table 2, which demonstrates the dominance of Hessian-based detectors, contradicting the results presented in Table 7.1 of [1].

*Conclusion:* It is demonstrated that the proposed repeatability criteria allow more reliable performance evaluation of feature detectors. With these criteria, it is found that Hessian-based detectors out-perform other state-of-the-art detectors on widely-used test datasets.

**Authors' affiliations:**
S. Ehsan, N. Kanwal, A. F. Clark and K. D. McDonald-Maier (School of Computer Science and Electronic Engineering, University of Essex, Colchester CO4 3SQ, United Kingdom)

**E-mail address of corresponding author:**
sehsan@essex.ac.uk

**Figure captions:**

Fig. 1   Repeatability scores and number of true matches for Hessian-Laplace detector with Bark dataset [5]

Fig. 2   Comparative analysis of state-of-the-art feature detectors for Bikes dataset [5] using Criterion 2

Fig. 1

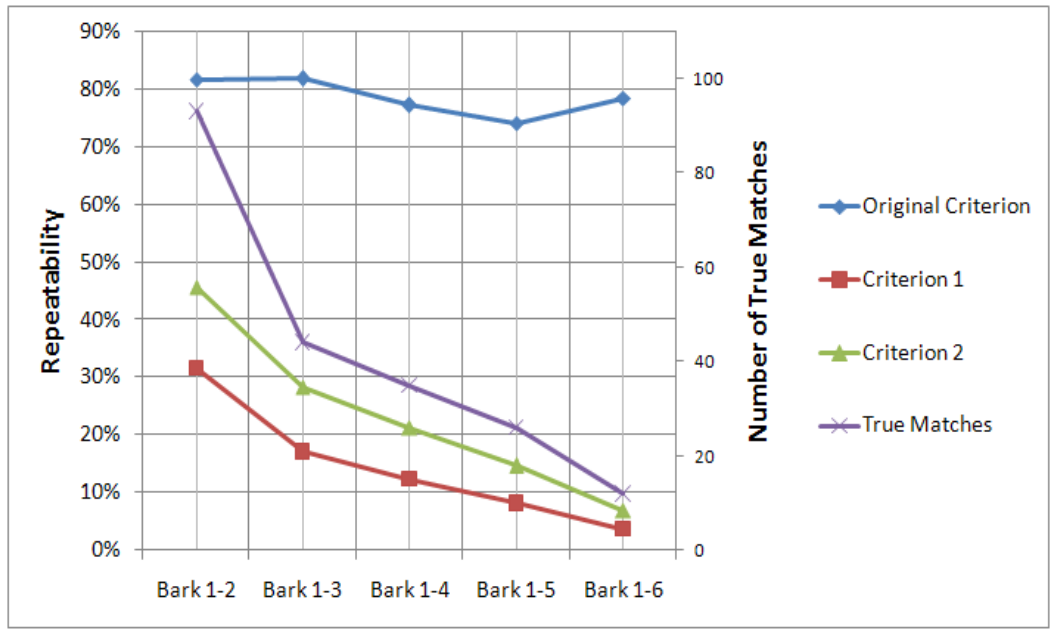

Fig. 2

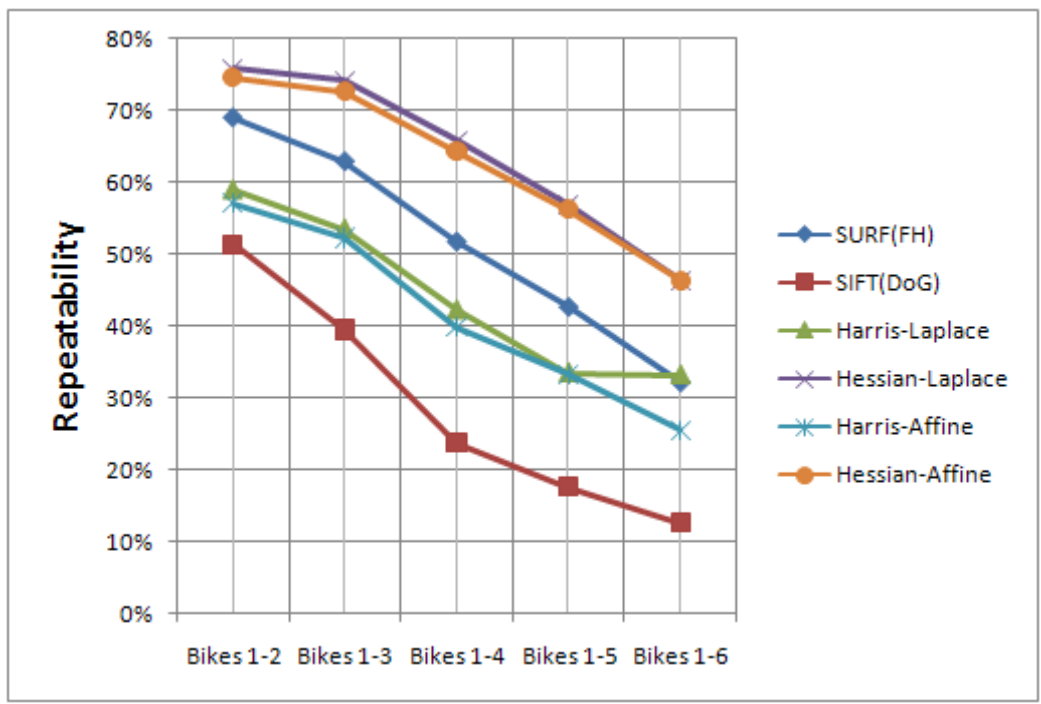

**Table captions:**

Table 1 Pearson's Correlation coefficients and corresponding p-values for repeatability curves with SURF detector for Oxford datasets [5]

Table 2 Summary of comparative analysis of six state-of-the-art feature detectors using Criterion 2

Table 1

| Image Datasets | Original Criterion | | Criterion 1 | | Criterion 2 | |
|---|---|---|---|---|---|---|
| | r | p-value | r | p-value | r | p-value |
| Bark | 0.884 | 0.0466 | 0.997 | 0.0002 | 0.989 | 0.0014 |
| Bikes | 0.7688 | 0.1287 | 0.996 | 0.0003 | 0.985 | 0.0022 |
| Boat | 0.697 | 0.1909 | 0.996 | 0.0003 | 0.993 | 0.0007 |
| Graffiti | 0.939 | 0.0179 | 0.971 | 0.0059 | 0.960 | 0.0095 |
| Leuven | 0.778 | 0.1213 | 0.998 | 0.0001 | 0.991 | 0.001 |
| Trees | -0.591 | 0.2940 | 0.991 | 0.001 | 0.968 | 0.0068 |
| UBC | 0.990 | 0.0012 | 0.998 | 0.0001 | 0.999 | 0.000 |
| Wall | 0.889 | 0.0436 | 0.950 | 0.0133 | 0.929 | 0.0225 |

Table 2

| Image Datasets | SIFT (DoG) | SURF (FH) | Harris-Laplace | Hessian-Laplace | Harris-Affine | Hessian-Affine |
|---|---|---|---|---|---|---|
| Bark | +++ | ++ | + | + | + | + |
| Bikes | + | +++ | ++ | +++ | ++ | +++ |
| Boat | ++ | ++ | ++ | +++ | + | ++ |
| Graffiti | + | + | + | + | +++ | +++ |
| Leuven | +++ | +++ | + | ++ | + | ++ |
| Trees | + | +++ | ++ | +++ | ++ | ++ |
| UBC | ++ | +++ | +++ | +++ | +++ | +++ |
| Wall | +++ | +++ | ++ | ++ | ++ | ++ |